\documentclass[journal]{IEEEtran}

\usepackage{epsfig} 
\usepackage[cmex10]{amsmath}

\usepackage{amssymb}  
\usepackage{bm} 
\usepackage{mathtools}

\usepackage[usenames,dvipsnames,svgnames]{xcolor}

\usepackage{booktabs} 
\usepackage{cite}
\usepackage{hyperref}

\makeatletter
\def\BState{\State\hskip-\ALG@thistlm}
\makeatother

\newtheorem{lemma}{Lemma}
\newtheorem{remark}{Remark}
\newtheorem{proposition}{Proposition}

\newcommand{\dif}{\mathrm{d}}
\newcommand{\e}{\mathrm{e}}

\newcommand{\lt}{\mathcal{L}}

\begin{document}
%
\title{Safety-Control of Mobile Robots Under Time-Delay Using Barrier Certificates and a Two-Layer Predictor}
%
%
%

\author{Azad Ghaffari and Manavendra Desai
\thanks{The authors are with the Department of Mechanical Engineering, Wayne State University, MI 48202, USA, {\tt\small aghaffari@wayne.edu}, {\tt\small manavendradesai@wayne.edu}.}}

\maketitle


\begin{abstract}
Performing swift and agile maneuvers is essential for the safe operation of autonomous mobile robots. Moreover, the presence of time-delay restricts the response time of the system and hinders the safety performance. Thus, this paper proposes a modular and scalable safety-control design that utilizes the Smith predictor and barrier certificates to safely and consistently avoid obstacles with different footprints. The proposed solution includes a two-layer predictor to compensate for the time-delay in the servo-system and angle control loops. The proposed predictor configuration dramatically improves the transient performance and reduces response time. Barrier certificates are used to determine the safe range of the robot's heading angle to avoid collisions. The proposed obstacle avoidance technique conveniently integrates with various trajectory tracking algorithms, which enhances design flexibility. The angle condition is adaptively calculated and corrects the robot's heading angle and angular velocity. Also, the proposed method accommodates multiple obstacles and decouples the control structure from the obstacles' shape, count, and distribution. The control structure has only eight tunable parameters facilitating control calibration and tuning in large systems of mobile robots. Extensive experimental results verify the effectiveness of the proposed safety-control.
\end{abstract}

\section{Introduction}

Obstacle avoidance has become an integral part of nonholonomic mobile robots, self-driving cars, unmanned aerial vehicles, and surface vehicles \cite{Peng2020, Liang2019, behjat2019learning, dong2020real, lim2020obstacle}. Prominent methods include potential field~\cite{Pan2019, karkoub2019trajectory}, collision cone~\cite{qu2004new, alonso2018cooperative}, path planning~\cite{fareh2020investigating, chu2012local}, receding horizon control~\cite{Pinkovich2020, Zhang2020, gao2008receding}, and fuzzy neural networks~\cite{kim2014obstacle}. Recently, barrier certificates and control barrier functions have gained attention to design safety-controllers~\cite{Kong2013, Romdlony2016, Ames2017, glotfelter2019hybrid}. 

However, analytical and numerical complexity restricts the practicality and scalability of existing safety methods for large-scale applications. For example, methods based on potential fields and barrier functions tie the safety-control law to the shape of the functions that model the obstacles. On the other hand, methods based on receding horizon optimization are numerically demanding. Thus, when the number of obstacles increases, one faces the cumbersome task of obstacle modeling and control implementation, which rapidly depletes processing resources. Also, simultaneous proof of stability and safety becomes an involved task.

Furthermore, time-delay in mobile robots causes long transients in trajectory tracking and delayed avoidance maneuvers which may cause collisions. In the presence of time-delay, any attempt to achieve fast transient by implementing high-gain controls often leads to oscillatory responses and instability. Hence, advanced techniques such as receding horizon optimal control~\cite{gao2008receding}, nonlinear predictor-based control~\cite{kojima2010predictor}, adaptive sliding mode control~\cite{roy2017adaptive}, and nonlinear tracking algorithm~\cite{park2017low} have been proposed to compensate time-delay for nonholonomic mobile robots.

Moreover, the Smith predictor is effective for time-delay compensation in linear time-invariant systems~\cite{molnar2019smith, ingimundarson2001robust, xing2018smith}. The nonholonomic mobile robot investigated in this work comprises DC motors and linear gear-boxes. Also, the wheel slip is negligible. Thus, one can accurately model the servo-system and heading angle using linear differential equations. Hence, a two-layer predictor, comprising three independent Smith predictors, is proposed to compensate for time-delay. First, the time-delay is compensated in the servo-system loops, which transfers the time-delay into the heading angle control. Second, another Smith predictor is embedded in the heading angle control to compensate for the transferred time-delay in steering the mobile robot.

Modular safety-control using barrier certificates has been proven effective to enforce safety nets around unmanned aerial vehicles~\cite{Ghaffari2018, Ghaffari2020}. Here, collision avoidance for mobile robots with time-delay is addressed such that control scalability is also achieved. Thus, the implementation and calibration effort remains minimal regardless of the number of robots and obstacles. Design objectives include high precision trajectory tracking, obstacle avoidance, and control scalability. Thus, a modular control structure is proposed. First, the two-layer predictor compensates time-delay and guarantees a desirable transient performance. The proposed predictor comprises an inner loop for the servo-system and an outer loop for angle tracking. Trajectory tracking is based on the vector-field-orientation (VFO) control, which guarantees accurate tracking by modifying the linear speed and heading angle~\cite{michalek2009vector}.

The safety-control is achieved by calculating the safe range of the heading angle using an exponential barrier certificate. Multiple obstacles with different shapes and arrangements can be modeled as a barrier certificate. The number and distribution of the obstacles only affect the barrier certificate's equation, and the rest of the algorithm remains unchanged. Each obstacle has an avoidance region with a tunable radius. The safety-control pushes the robot away from the avoidance region as soon as the actual heading angle projects a collision with the obstacle. The safety condition provides an inequality that ties the position, safe heading angle, and translational velocity of the robot to the obstacle's location and avoidance radius. The proposed safety algorithm provides a dynamic estimate of the safe heading angle. Moreover, when the robot is far enough from the obstacle, the safety algorithm is inactive.

The proposed modular safety-control dramatically improves control scalability by utilizing linear control components, Smith predictor, and safe heading angle estimate. Regardless of the number, shape, and distribution of the obstacles or the reference trajectory, the proposed algorithm maintains safe trajectory tracking for the nonholonomic mobile robot. Also, the algorithm's analytical complexity and needed processing power are independent of the number of obstacles. The proposed design is flexible and easy to tune and calibrate. Therefore, improved scalability is a byproduct of the proposed algorithm.

This paper is built upon the work by Ghaffari~\cite{Ghaffari2021}. This work presents three main contributions: 1) the two-layer predictor to compensate time-delay, 2) embedded modular safety-control structure, and 3) extensive experimental results to verify the effectiveness of the proposed algorithm when facing obstacles of different size and shape. Tracking accuracy is achieved as high as 98\%, and the response time is improved by a factor of four. Hence, unlike the preliminary work~\cite{Ghaffari2021}, this paper does not utilize the contour error estimate and feedback modification to improve trajectory tracking precision. The two-layer predictor and the experimental results are reported here for the first time.

The rest of this paper is presented in the following order. Dynamic model and trajectory tracking using the VFO algorithm is explained in Section~\ref{sec:model}. The two-layer predictor is described in Section~\ref{sec:SP}. The safety-control and allowable heading angle are presented in Section~\ref{sec:safety}. Section~\ref{sec:sim} presents experimental results to verify and validate the effectiveness of the proposed method. Section~\ref{sec:con} concludes the paper.

\section{Dynamic Model and Trajectory Tracking}\label{sec:model}

\begin{figure}
\begin{center}
\includegraphics[clip,width=0.5\columnwidth]{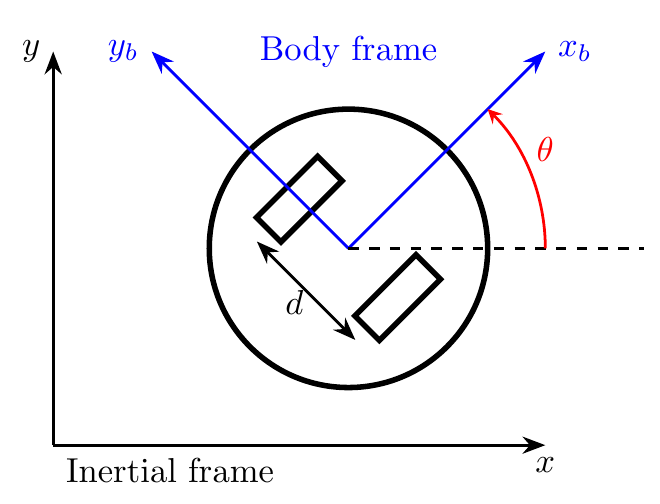}
\end{center}
\vspace{-5mm}
\caption{A schematic of a nonholonomic mobile robot in an inertial reference frame $x$-$y$. The body-fixed reference frame is $x_b$-$y_b$ and $\theta$ is the heading angle.}
\label{fig:nmr}
\end{figure}

The nonholonomic mobile robot is driven by a differential-drive system comprised of two identical electric wheels. Fig.~\ref{fig:nmr} shows the schematic of the robot and inertial and body-fixed reference frames. The body-fixed reference frame is attached to the robot at the center of mass. The heading angle is $\theta$, and is measured with respect to the $x$-axis. The kinematic equations of the robot are given as
\setlength{\arraycolsep}{1pt}%
\begin{eqnarray}
\label{eq:dx}
\dot{x}&=&v\cos\theta\\
\label{eq:dy}
\dot{y}&=&v\sin\theta\\
\label{eq:dtheta}
\dot\theta&=&\omega,
\end{eqnarray}
where $[x~~y]^ T $ is the robot's position in the inertial reference frame, $ \theta $ is the heading angle, $v$ is linear velocity, and $\omega$ is angular velocity. 

The force and torque generated by the electric wheels control the linear and angular velocity of the robot, respectively. Assume that the effect of friction and wheel slip on the dynamic model is negligible, and the left and right wheel's movements do not affect each other. Thus, one can decouple the dynamic equation of the left and right wheels.  Since the mobile robot is symmetric with identical left and right wheels and motors, it is reasonable to assume that both wheels have the same dynamic equation given as
\begin{eqnarray}
\label{eq:Gwheel}
\frac{V_i(s)}{U_i(s)}&=&G(s)\e^{-\tau s}
\end{eqnarray}
where $G(s)$ is the a transfer function, $\tau$ is the constant time-delay, $V_i(s)={\lt}(v_i)$, and $U_i(s)={\lt}(u_i)$, where ${\lt}$ denotes the Laplace transform. The wheel linear velocity is $v_i$ and the wheel voltage is $u_i$, where $i=R,L$ for the right and left wheel, respectively. The voltage of the electric wheels are bounded as $|u_i|\le u_\text{max}$ for $i=R, L$, where $u_\text{max}$ is a positive constant. The robot's linear and angular velocity are related to the wheels' velocity through the following equations
\begin{eqnarray}
\label{eq:vu}
v&=&\big(v_R+v_L\big)/2\\
\label{eq:omegau}
\omega&=&\big(v_R-v_L\big)/d,
\end{eqnarray}
where $d$ is the distance between the center of the wheels. Therefore, the equations of linear and angular velocity can be written as the following
\begin{eqnarray}
\label{eq:Gv}
\frac{V(s)}{W_i(s)}&=&G(s)\e^{-\tau s},
\end{eqnarray}
where $V(s)={\lt}(v)$, $\Omega(s)=\lt(\omega)$, $W_i(s)={\lt}(w_i)$ for $i=v, \omega$, where 
\begin{eqnarray}
\label{eq:wvu}
w_v&=&\big(u_R+u_L\big)/2\\
\label{eq:womegau}
w_\omega&=&\big(u_R-u_L\big)/d.
\end{eqnarray}


The vector-field-orientation (VFO) method proposed by Michalek and Kozlowski~ \cite{michalek2009vector} is utilized to determine adjusted linear and angular velocities, $v_a$ and $\omega_a$, for the servo-system (inner loop) of the mobile robot. Given the instantaneous position $[x~~y]^T$ and heading angle $\theta$ of the mobile robot, the essence of the VFO lies in \textit{orienting} the mobile robot along an adjusted direction $\theta_a$, and \textit{pushing} it along this direction with adjusted velocity $v_a$, to eventually reach the reference position $[x_r~~y_r]^T$ with a reference orientation $\theta_r$. Thus, the differential-drive nature of the mobile robot model is effectively reduced to that of a unicycle.

The reference trajectory is constructed such that the nonholonomic condition is satisfied
\begin{eqnarray}
\label{eq:dxr}
\dot{x}_r&=&v_r\cos\theta_r\\
\label{eq:dyr}
\dot{y}_r&=&v_r\sin\theta_r\\
\label{eq:dthetar}
\dot\theta_r&=&\omega_r.
\end{eqnarray}
The proper selection of $v_r$ and $\omega_r$ creates a variety of reference trajectories. Also, this paper only considers forward movement. Thus, $v_r$ is a positive real value. However, $\omega_r$ can take a positive or negative real value. The VFO control is used for trajectory tracking control. The algorithm calculates the adjusted linear velocity, $v_a$, and heading angle, $\theta_a$, from the position error. 

Denote the position error variables as $e_x(t)=x_r(t)-x(t)$ and $e_y(t)=y_r(t)-y(t)$. The adjusted linear velocity and heading angle are obtained as the following
\begin{eqnarray}
v_a(t)&=&h_x(t)\cos\theta+h_y(t)\sin\theta\\
\label{eq:thetaA}
\theta_a(t)&=&\text{atan2c}\big(h_y(t),h_x(t)\big),
\end{eqnarray}
where
\begin{eqnarray}
h_x(t)&=&ke_x(t)+\dot{x}_r(t)\\
h_y(t)&=&ke_y(t)+\dot{y}_r(t),
\end{eqnarray}
where $k>0$, and $h(t)=[h_x(t)~~h_y(t)]^T$ is the \textit{convergence vector field}, which can be interpreted as the desired linear velocity of the mobile robot. Note that $q(\theta)=[\cos\theta~~\sin\theta]^T$ interprets the instantaneous heading of the mobile robot. Thus, the adjusted linear velocity $v_a(t)$ is defined as the projection of $h(t)$ on $q(\theta)$. This ensures that the mobile robot is pushed only in proportion to the extent of collinearity between $\theta$ and $\theta_a(t)$. 
Moreover, the function $\text{atan2c}\left(a,b\right)$ is the four quadrant arctangent of $a$ and $b$, which is implemented such that $\theta_a(t)$ provides a continuous and differentiable curve and is not wrapped between $[-\pi,\pi]$. For more information please see the work of Michalek and Kozlowski~\cite{michalek2009vector}. 

For situations where the time-delay is negligible, one can obtain the adjusted angular velocity as the following
\begin{eqnarray}
\label{eq:wa}
\omega_a&=&K_{P,\theta}e_\theta(t)+K_{I,\theta}\int_0^te_\theta(\eta)\dif\eta+\dot\theta_a(t),
\end{eqnarray}
where $K_{P,\theta}>0$ and $k_{I,\theta}>0$ are control gains, $e_\theta(t)=\theta_a(t)-\theta(t)$, and 
\begin{equation}
\dot\theta_a(t)=\frac{\dot{h}_y(t)h_x(t)-h_y(t)\dot{h}_x(t)}{h_x^2(t)+h_y^2(t)},
\end{equation}
where
\begin{eqnarray}
\dot{h}_x(t)&=&k\Big(\dot{x}_r(t)-v_a\cos\theta\Big)+\ddot{x}_r(t)\\
\dot{h}_y(t)&=&k\Big(\dot{y}_r(t)-v_a\sin\theta\Big)+\ddot{y}_r(t).
\end{eqnarray}
The integral action is added to \eqref{eq:wa} to further reduce the tracking error in steady-state. Using \eqref{eq:vu} and \eqref{eq:omegau}, the adjusted wheel velocity values for the servo-system control loop are found using the following relationships
\begin{eqnarray}
\label{eq:vRa}
v_{R,a}&=&v_a+\frac{d}{2}\omega_a\\
\label{eq:vLa}
v_{L,a}&=&v_a-\frac{d}{2}\omega_a.
\end{eqnarray}
Since the wheel voltage is limited, the obtained values of $v_{R,a}$ and $v_{L,a}$ are scaled to avoid control saturation 
\begin{eqnarray}
v_{i,sc}=\left\{\begin{array}{lcl}{v_{i,a}}/{\mu} & ~~\text{if}~~ & \mu>1\\ v_{i,a} & ~~\text{if}~~ & \text{otherwise}\end{array}\right., \quad i=R, L,
\end{eqnarray}  
where $\mu=\max\left(|u_R|,|u_L|\right)/{u_\text{max}}$, and $v_{i,sc}$ is the scaled wheel voltage, for $i=R,L$. 

The success of the VFO heavily depends on the performance of the angle tracking and servo-system. If the servo-system and angle tracking are not enough fast and accurate, the VFO performance deteriorates, and the trajectory tracking may fail. Since time-delay limits the system's response time, one needs to compensate for the effect of time-delay to achieve a desirable trajectory tracking performance. Thus, the next section presents a two-layer predictor that guarantees enough fast transient responses.

\section{Two-Layer Predictor}\label{sec:SP}

\begin{figure*}
\begin{center}
\includegraphics[width=0.7\textwidth]{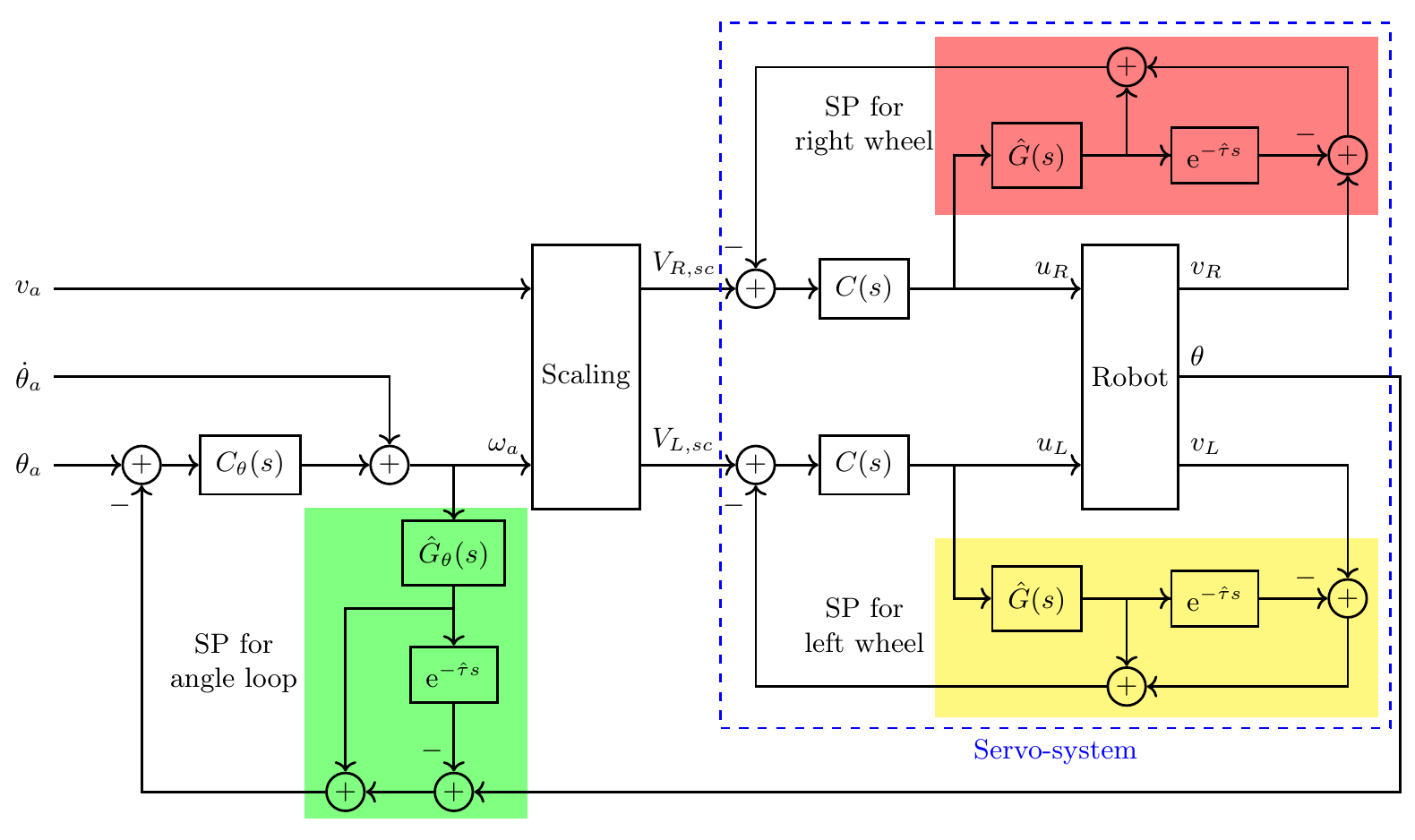}
\end{center}
\vspace{-5mm}
\caption{Block diagram of the proposed two-layer predictor. The innermost loops (highlighted in red and yellow) compensate for the time-delay in the servo-system, and the outer-loop (highlighted in green) compensates the transferred time-delay to the heading angle control from the servo-system.}
\label{fig:SPblockdiagram}
\end{figure*}
The robot is driven using DC motors and linear gear-boxes, and the wheel slip is negligible. Moreover, reasons such as communication lags and actuator or sensor properties may cause time-delay in the servo-system. Here, a constant time-delay is considered in the input of the servo-system. Thus, each wheel is modeled as~\eqref{eq:Gwheel}, which is a combination of a transfer function and constant time-delay.  If the time-delay is comparable to the response time of the servo-system, one needs to compensate for the effect of time-delay to improve the transient performance of the system. 

As shown in Fig.~\ref{fig:SPblockdiagram}, the Smith predictor (SP) compensates time-delay by acting on a nominal model of the system to provide a controlled response that is unaffected by time-delay. Furthermore, the Smith predictor compares the actual system output to the nominal delayed-output to eliminate drifts and external disturbances in the system response. Since heading angle determines the robot's translational motion, a layer of predictor only in the servo-system won't achieve the best tracking precision. Thus, another layer of predictor is considered for the heading angle. The two layers shown in Fig.~\ref{fig:SPblockdiagram} work in tandem to achieve a desirable transient performance for trajectory tracking.

Since the servo-system of the wheels are assumed identical, the velocity control for the two wheels are also identical. Consider the nominal model of each wheel is given as
\begin{equation}
\label{eq:hG}
\frac{\hat{V}_i(s)}{U_i(s)}=\hat{G}(s)\e^{-\hat\tau s}, \quad i=R, L,
\end{equation} 
where $\hat{V}_i$ is the estimate of wheel velocity, and $\hat{G}(s)$ and $\hat\tau$ are found using mathematical modeling or system identification, which might be different from the actual values $G(s)$ and $\tau$. One can use \eqref{eq:hG} to obtain an estimate of the wheel velocity, $\hat{V}_i(s)=\hat{G}(s)\e^{-\hat\tau s}U_i(s)$, where $U_i(s)$ is the input voltage of the wheel, where $i=R, L$. Moreover, one can predict the future output as $\hat{V}_i(s)\e^{\hat\tau s}$. Thus, the feedback signal can be corrected as the following
\begin{equation}
\label{eq:control}
U_i(s)=C(s)\left(V_{i,sc}(s)-\left(V_i(s)-\hat{V}_i(s)+\hat{V}_i(s)\e^{\hat\tau s}\right)\right),
\end{equation} 
where $C(s)$ is the control and $V_{i,sc}$ is the scaled adjusted velocity obtained from the VFO, where $i=R,L$. Using \eqref{eq:hG}, an implementable realization of the \eqref{eq:control} is obtained as
\begin{equation}
U_i(s)=C(s)\bigg(V_{i,sc}(s)-\Big(V_i(s)+Z(s)U_i(s)\Big)\bigg), i=R, L,
\end{equation}
where 
\begin{equation}
Z(s)=\hat{G}(s)-\hat{G}(s)\e^{-\hat\tau s}.
\end{equation}
The controller with the Smith predictor can be described as
\begin{equation}
C_{sp}(s)=\frac{C(s)}{1+C(s)Z(s)}.
\end{equation}
Hence, the closed-loop transfer function of the servo-system becomes
\begin{eqnarray}
\hspace{-5mm}G_{cl}&=&\frac{C_{sp}(s)G(s)\e^{-\tau s}}{1+C_{sp}(s)G(s)\e^{-\tau s}}\nonumber\\
&=&\frac{C(s)G(s)\e^{-\tau s}}{1\!+\!C(s)\hat{G}(s)\!-\!C(s)\hat{G}(s)\e^{-\hat\tau s}+C(s)G(s)\e^{-\tau s}}.
\end{eqnarray}

For the ideal case where the nominal model and delay are perfectly known, i.e., $\hat{G}(s)=G(s)$ and $\hat\tau=\tau$, the closed-loop transfer function of the servo-system simplifies to
\begin{equation}
\label{eq:vcl}
\frac{V_i(s)}{V_{i,sc}}=G_{v,cl}(s)\e^{-\tau s},
\end{equation}
where 
\begin{equation}
G_{v,cl}(s)=\frac{C(s)G(s)}{1+C(s)G(s)}
\end{equation}
Therefore, the Smith predictor removes the effect of time-delay on the closed-loop poles of the servo-system and moves the time-delay to outside of the servo-system loop. For a detailed look inside the Smith predictor, please see~\cite{molnar2019smith, ingimundarson2001robust}.

The first layer of the Smith predictor transfers time-delay to angle and position loops. Achieving proper orientation is key for the success of the VFO. Thus, the effect of transferred time-delay on the angle control needs to be compensated. Assume that the wheel voltages are not saturated, $v_{i,sc}=v_{i,a}$ for $i=R,L$. Using \eqref{eq:omegau} and \eqref{eq:vcl}, one obtains
\begin{equation}
\label{eq:Gwcl}
\frac{\Omega(s)}{\Omega_a(s)}= G_{v,cl}\e^{-\tau s},
\end{equation}
where $\Omega_a(s)={\lt}(\omega_a)$, where \eqref{eq:vRa} and \eqref{eq:vLa} give  
\begin{equation}
\omega_a=\frac{v_{R,a}-v_{L,a}}{d}.
\end{equation}
As mentioned earlier, the trajectory tracking performance depends on the servo-system and the angle control to a more considerable extent. Furthermore, the VFO convergence rate is limited by the servo-system and angle control convergence rate. In other words, the inner-loop control must achieve an adequately fast transient response in comparison with the VFO. Hence, another Smith predictor is designed for the angle control loop. 

Using equation \eqref{eq:dtheta} and \eqref{eq:Gwcl}, one obtains
\begin{equation}
\frac{\Theta(s)}{\Omega_a(s)}=G_{\theta}\e^{-\tau s},\quad G_{\theta}={G_{v,cl}}/{s},
\end{equation}
where $\Theta(s)={\lt}(\theta(t))$.
Thus, the angle control with the Smith predictor correction is implemented as the following
\begin{equation}
\Omega_a(s)=C_\theta(s)\Big(\Theta_{a}(s)-\big(\Theta(s)+Z_\theta(s)\Omega_a(s)\big)\Big),
\end{equation}
where 
\begin{equation}
Z_\theta(s)=\hat{G}_\theta(s)-\hat{G}_\theta(s)\e^{-\hat\tau s},
\end{equation}
where $\hat{G}_\theta(s)=\hat{G}_{v,cl}/s$, where 
\begin{equation}
\hat{G}_{v,cl}(s)=\frac{C(s)\hat{G}(s)}{1+C(s)\hat{G}(s)}.
\end{equation}
The block diagram of the two-layer predictor is shown in Fig.~\ref{fig:SPblockdiagram}. The adjusted values of, $v_a$, $\theta_a$, and $\dot\theta_a$, are produced by the VFO control. The angle control and servo-system control is shown as $C_\theta(s)$ and $C(s)$, respectively.

\begin{remark}
The time-delay determines the response time of the servo-system and the angle control. Moreover, the VFO is required to act slower than the inner-loops, including the angle control and the servo-system. In other words, the inner-loops need to settle down long before the VFO settles down. Thus, the response time of the VFO will be enough larger than the sample-time. Therefore, an additional predictor for the VFO may neither be an appropriate design nor can lead to a noticeable improvement in trajectory tracking performance.
\end{remark}

\section{Safe Heading Angle and Collision Avoidance}\label{sec:safety}

Popular obstacle avoidance control techniques such as artificial potential fields, collision cones, and receding horizon optimization tie the control law to the obstacle properties. Hence, any changes in the number, shape, and distribution of the obstacles lead to control recalculation, which complicates control design and prolongs the control calibration. Thus, in this paper, exponential barrier certificates are used to isolate the safety-control from obstacle properties. The proposed algorithm does not interfere with the trajectory tracking control, simplifies the collision avoidance control design, and minimizes the processing power required to implement the safety-control. 

Consider the following system 
\begin{equation}\label{eq:general}
\dot{X}=F(X), \quad X\in\mathcal{X}\subseteq\mathbb{R}^n,
\end{equation}
where $F(X)$ is smooth enough. A set of initial conditions $\mathcal{X}_0\in\mathcal{X}$  and a set of unsafe states $\mathcal{X}_u\subset\mathcal{X}$ are given. The safety is achieved if all the state trajectories initiated inside $\mathcal{X}_0$ avoid the unsafe set for all $t>0$. The following lemma is introduced to use barrier certificates to obtain the safe limits of the robot's heading angle.

\begin{lemma}[Exponential Safety Condition~\cite{Kong2013}] \label{Lemma1} 
Consider the system~\eqref{eq:general} and the corresponding sets $\mathcal{X}, \mathcal{X}_0,$ and $\mathcal{X}_u$. For any given $\alpha\in\mathbb{R}$, if there exists a barrier certificate, i.e., a continuously differentiable function $\mathcal{B}(X): \mathcal{X}\rightarrow\mathbb{R}$ satisfying the following conditions:
\begin{eqnarray}
\label{eq:B1}
\forall X\in\mathcal{X}_0:&& \mathcal{B}(X)\le 0 \\
\label{eq:B2}
\forall X\in\mathcal{X}_u: && \mathcal{B}(X)>0\\
\label{eq:B3}
\forall X\in\mathcal{X}: && \left({\partial \mathcal{B}(X)}/{\partial X}\right)f(X)\le -\alpha \mathcal{B}(X)
\end{eqnarray}
then the safety property is satisfied by the system~\eqref{eq:general}, i.e., $\mathcal{B}(X(t))\le0$ for all $t>0$.
\end{lemma}

Note that $\mathcal{B}(X)=0$ shows the avoidance boundary. Also, for $\alpha>0$, the system can be steered very close to the avoidance boundary without violating the safety condition. A negative or zero value of $ \alpha $ creates a repelling avoidance boundary leading to a conservative control design, which is not suitable to augment with the trajectory tracking control. Thus, throughout this paper, positive values of $\alpha$ are considered.

Consider an avoidance boundary modeled as a barrier certificate function $B(x,y)$ satisfying conditions~\eqref{eq:B1}--\eqref{eq:B3}. Note that the barrier certificate's choice is optional, and multiple obstacles with different shapes could be modeled using a single barrier certificate. Denote the gradient and Hessian of the barrier certificate as
\begin{eqnarray}
g&=&\left[\frac{\partial B(x,y)}{\partial x}~~\frac{\partial B(x,y)}{\partial y}\right]\\
H&=&\left[\begin{array}{cc}\frac{\partial^2 B(x,y)}{\partial x^2} & \frac{\partial^2 B(x,y)}{\partial x\partial y}\\\frac{\partial^2 B(x,y)}{\partial x\partial y}& \frac{\partial^2 B(x,y)}{\partial y^2}\end{array}\right].
\end{eqnarray}

Using the kinematic equations~\eqref{eq:dx} and \eqref{eq:dy}, the following is obtained using~\eqref{eq:B3}
\begin{eqnarray}
g_1 \dot{x}+g_2\dot{y}\le -\alpha B(x,y)\\
\label{ineq:B1}
g_1v\cos\theta+g_2v\sin\theta\le -\alpha B(x,y).
\end{eqnarray}
Since $v>0$, inequality~\eqref{ineq:B1} can be written as
\begin{eqnarray}
\label{ineq:B2}
g_1\cos\theta+g_2\sin\theta &\le& -\alpha \frac{B(x,y)}{v}\\
\label{ineq:B3}
\frac{g_1}{\|g\|}\cos\theta+\frac{g_2}{\|g\|}\sin\theta&<&c,
\end{eqnarray}
where $c=-\alpha{B(x,y)}/{(v\|g\|)}$, and $\|\cdot\|$ denotes the Euclidean norm. 

Denote 
\begin{eqnarray}
\label{eq:beta1}
\cos\beta=\frac{g_1}{\|g\|}\\
\label{eq:beta2}
\sin\beta=\frac{g_2}{\|g\|}.
\end{eqnarray}
Thus, inequality~\eqref{ineq:B3} gives the angle condition
\begin{equation}
\label{ineq:angle}
\cos(\theta-\beta)\le c.
\end{equation}
Note that $c>0$ outside the avoidance region because $v>0$ and $B(x,y)<0$. Moreover, if $c>1$, the angle condition \eqref{ineq:angle} gives
\begin{equation}
\cos\left(\theta-\beta\right)\le1,
\end{equation}
which is true for any value of $\theta$. Thus, the robot heading angle is not restricted, and the robot can safely track the reference trajectory. On the other hand, if $c<1$, there is a range of safe heading angles prohibiting the robot from entering the avoidance region. 

To calculate the safe limits of the heading angle, first, angle $\delta$ is introduced such that
\begin{equation}
\cos\delta=c.
\end{equation}
Recall that $c>0$ outside the avoidance region. Thus, it is obtained that $0\le\delta\le\pi/2$. Hence, one can calculate $\delta$ as shown in the following
\begin{equation}
\delta=\arccos\left(c\right).
\end{equation}
Therefore, the angel condition~\eqref{ineq:angle} gives
\begin{equation}
\label{eq:cos}
\cos\left(\theta-\beta\right)\le\cos\delta
\end{equation}
The safety result about the heading angle is summarized in the following proposition.
\begin{proposition}
Consider the kinematic equation of the mobile robot is given as~\eqref{eq:dx}--\eqref{eq:dtheta}, where the linear velocity is positive. Assume obstacles are described as a barrier certificate $B(x,y)$, where $B(x,y)>0$ inside the avoidance zone of obstacles, and $B(x_0,y_0)\le0$, where $[x_0~~y_0]^T$ is the initial position of the mobile robot. There exist a positive $\alpha$ and a set of angular velocities such that the heading angle satisfies \eqref{eq:cos} for all $t>0$. Then, the mobile robot does not enter the avoidance zone of any obstacle.
\end{proposition}

Condition~\eqref{eq:cos} gives the unsafe range of heading angle as $2k'\pi-\delta\le\theta-\beta\le2k'\pi+\delta$ for $k'=0,\pm1,\pm2,\cdots$. However, the heading angle cannot change abruptly. Therefore, one can neglect non-zero values of $k'$, and express the unsafe range of the heading angle as the following set
\begin{eqnarray}
\label{eq:Xu}
\Theta_u&=&\left\{\theta\in\mathbb{R}: -\delta\le\theta-\beta\le\delta\right\}\nonumber\\
\label{eq:Theta}
&=&\left\{\theta\in\mathbb{R}: \beta-\delta\le\theta\le \beta+\delta\right\}
\end{eqnarray}

Since $\beta$ can take any real values, applying the inverse trigonometric functions to~\eqref{eq:beta1} and \eqref{eq:beta2} is not a viable method to calculate $\beta$ in real-world applications. Here, a dynamic update law is provided to ensure robust calculation. Taking the derivative of both sides of $\tan\beta=g_2/g_1$ gives
\begin{eqnarray}
\dot\beta\left(1+\tan^2\beta\right)&=&\frac{\dot{g}_2g_1-\dot{g}_1g_2}{g_1^2}\\
\dot\beta\left(1+\frac{g_2^2}{g_1^2}\right)&=&\frac{\dot{g_2}g_1-\dot{g_1}g_2}{g_1^2}\\
\label{eq:dbeta1}
\dot\beta&=&\frac{\dot{g}_2g_1-\dot{g}_1g_2}{\|g\|^2}.
\end{eqnarray}
The barrier certificate function only depends on the robot's position. Thus, $g_1$ and $g_2$ depend on $x$ and $y$. Also, recall that $\dot{x}$ and $\dot{y}$ are expressed as \eqref{eq:dx} and \eqref{eq:dy}, respectively. Hence, $\dot{g}_1$ and $\dot{g}_2$ depend on $x,~y,~v,$ and $\theta$. Therefore, it is obtained that $\dot{g}_1=\phi_1(x,y,v,\theta)$ and $\dot{g}_2=\phi_2(x,y,v,\theta)$, where
\begin{eqnarray}
\phi_1(x,y,v,\theta)&=&H_{11}v\cos\theta+H_{12}v\sin\theta\\
\phi_2(x,y,v,\theta)&=&H_{21}v\cos\theta+H_{22}v\sin\theta
\end{eqnarray}
Thus, \eqref{eq:dbeta1} can be rewritten as
\begin{equation}
\dot\beta=\frac{v\phi(x,y,v,\theta)}{\|g\|^2},
\end{equation}
where 
\begin{eqnarray}
\phi(x,y,v,\theta)&=&g_1(H_{21}\cos\theta+H_{22}\sin\theta)-\nonumber\\
&&{}-g_2(H_{11}\cos\theta+H_{12}\sin\theta).
\end{eqnarray}
Recall that $\beta$ is calculated for $c<1$. Thus, the initial value of $\beta$ is reset to the value of the heading angle at the moment where $c<1$ for the first time.

Since the heading angle must avoid the set $\Theta_u$, the adjusted angle, $\theta_a$, obtained from the VFO is further modified to avoid the unsafe set. The robot can either turn left or right to avoid the obstacles. Here, left turn avoidance maneuver is considered. Thus, the $\theta_a$ is modified as the following
\begin{eqnarray}
\label{eq:leftturn}
\theta_s=\left\{\begin{array}{lcl}\theta_a &~~\text{if}~~& \theta_a\notin\Theta_u\\ \beta+\delta &~~\text{if}~~& \theta_a\in\Theta_u\end{array}\right.,
\end{eqnarray}
where $\theta_s$ is the safe heading angle.
Right turn avoidance maneuver is obtained using the following logic
\begin{eqnarray}
\label{eq:rightturn}
\theta_s=\left\{\begin{array}{lcl}\theta_a &~~\text{if}~~& \theta_a\notin\Theta_u\\ \beta-\delta &~~\text{if}~~& \theta_a\in\Theta_u\end{array}\right..
\end{eqnarray}

The low and high limits of the unsafe heading angle change depending on the robot's location and translational velocity. Thus, the values of $\beta$ and $\delta$ are calculated in real-time. Moreover, the angle control requires the derivative of the safe angle. A high-pass filter is utilized to create an estimate of $\dot\theta_s$
\begin{equation}
z=\frac{s}{Ts+1}\theta_s,
\end{equation}
where $T$ is small enough in comparison to the angle control response time.

If the robot's heading angle falls inside the unsafe angle range, the safety control is activated. Then, the safe angle, $\theta_s$, and its estimated time derivative, $z$, replace the adjusted heading angle, $\theta_a$, and its time derivative, $\dot\theta_a$. Therefore, the adjusted rotational speed, $\omega_a$, is accordingly modified. However, the adjusted linear velocity, $v_a$, is calculated using the position error. The produced values by the VFO for $v_a$ are not reliable during avoidance maneuver. Thus, the adjusted linear velocity is replaced by the reference linear velocity, $v_r$, during the avoidance maneuver, i.e.,
\begin{eqnarray}
v_s=\left\{\begin{array}{lcl}v_a &~~\text{if}~~& \theta_a\notin\Theta_u\\ v_r &~~\text{if}~~& \theta_a\in\Theta_u\end{array}\right.,
\end{eqnarray} 

The details of the conducted experiments are explained in the next section. It is shown that the combination of the VFO, two-layer predictor, and safety-control can achieve adequately precise trajectory tracking performance with guaranteed safety behavior.

\section{Experimental Results}\label{sec:sim}

\begin{figure}
\begin{center}
{\bf (a)}\\
\includegraphics[width=0.8\columnwidth]{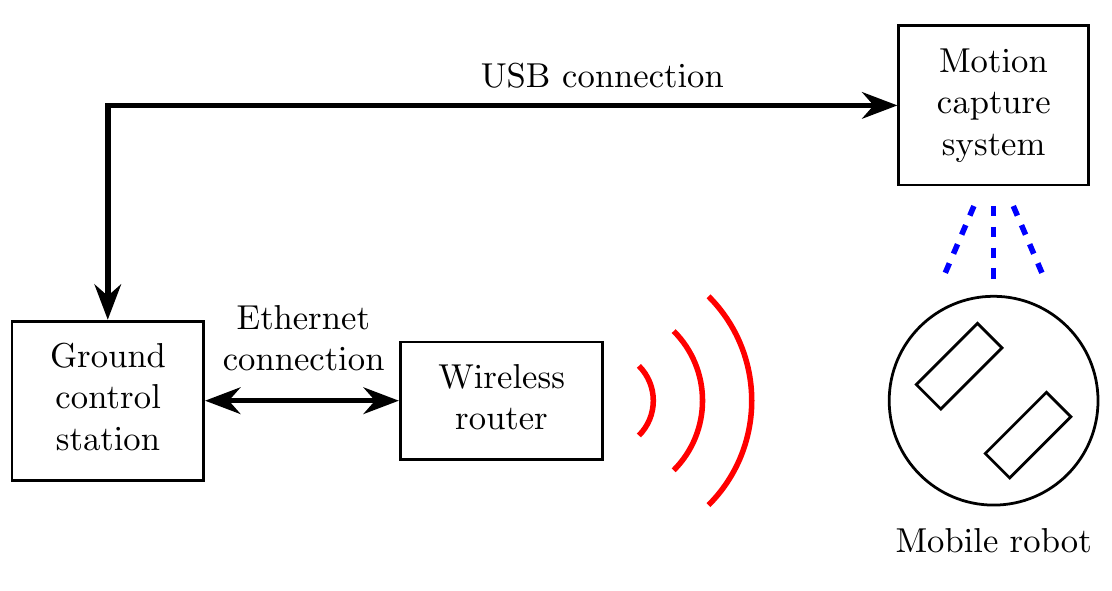}\\
{\bf (b)}\\
\includegraphics[width=0.3\columnwidth]{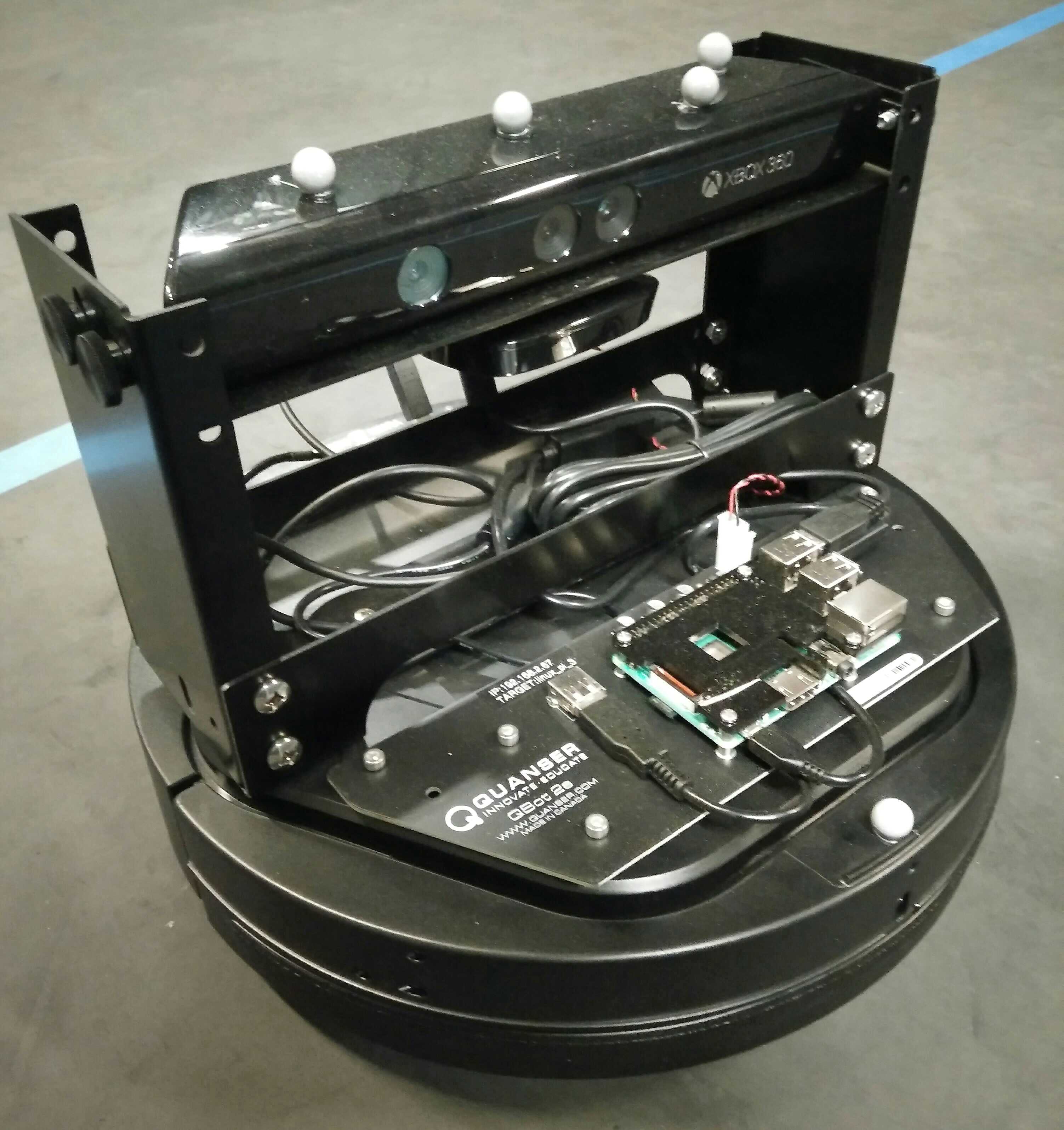}
\end{center}
\vspace{-5mm}
\caption{(a) Components of the experimental testbed. (b) Quanser's QBot~2e used in the experiments~\cite{QuanserQBot2e2019}. The white spheres are passive markers.}
\label{fig:AVRS}
\end{figure}
\begin{table}
\caption{Technical parameters of the mobile robot~~\cite{QuanserQBot2e2019}}
\vspace{-5mm}
\label{tab:robot}
\begin{center}
\begin{tabular}{ll}
\toprule
Parameter & Value\\
\midrule
Diameter & $35$~cm\\
Total mass & $3.82$ kg\\
Maximum translational velocity & $70$ cm/s\\
Maximum rotational velocity & $180$~$^\circ$/s\\
Distance between right and left wheels & $23.50$ cm\\
Encoder & $11.73$ ticks/mm\\
\bottomrule
\end{tabular}
\end{center}
\end{table}
Extensive experiments are carried out to verify the effectiveness of the proposed safety-control. As shown in Fig.~\ref{fig:AVRS}, the experimental testbed includes a ground control station, which is used to code, compile, and download the executable files to the mobile robot. The ground control station also acts as a data acquisition system. The position and orientation information is acquired using a motion capture system, which comprises eight Flex~13 infrared cameras. The cameras are connected to the ground control station through two USB hubs. A wireless router is used to communicate with the mobile robot. The mobile robot is equipped with a processing board that allows running the control loops locally. The servo-system of each wheel is comprised of a DC motor and a gearbox, which amplifies the produced torque of the DC motor. The velocity of each wheel is numerically derived from the angular positions of the respective axle, that are measured using encoders. Technical information of the mobile robot is given in Table~\ref{tab:robot}.

As shown in Fig.~\ref{fig:AVRS}(b), the robot has six passive markers which allow the motion capture system to measure the robot's position and orientation in the operating environment. The linear and rotational velocities are produced using the wheel velocities and from \eqref{eq:vu} and \eqref{eq:omegau}. The control sample rate is set to 1~ms throughout this paper.

System identification is carried out to obtain the dynamic equation of the vehicle. The results show that the wheels behave independently and a linear model with time-delay can model each wheel. A quantitative system identification shows that a second-order transfer function with input delay fits the estimation data with accuracy above $85\%$. The wheel model from the DC motor voltage to the wheel linear velocity is obtained as
\begin{equation}
\frac{V_i(s)}{U_i(s)}=\frac{5.94s+1.45}{s^2+7.40s+1.42}\e^{-0.50s}, \quad i=R,L,
\end{equation}
where the input and output units are V and m/s, respectively.

The wheel dynamic has a pole at $s=-0.20$ and a zero at $s=-0.24$, which causes a lengthy transient. Moreover, time-delay restricts a controller's ability to reduce the response time of the system to a desirable level. Numerical simulations were conducted a priori to design the control and initialize the control parameters, evaluate the proposed safety-control capabilities, and troubleshoot the implementation issues. In general, the conducted numerical simulations align with the experiments. However, modeling error and system uncertainty cause a noticeable deviation between numerical simulations and experimental results. Hence, the experimental setup is used to calibrate the control parameters. The subsequent analysis only reports the experimental results. 

The experiments are presented as follows: 1) the effect of the two-layer predictor on the trajectory tracking is investigated, 2) the safety algorithm is tested with two circular obstacles in the operating environment, and 3) the obstacle avoidance maneuver is tested for a large obstacle with non-circular avoidance zone in the operating environment.  
 
\subsection{First Experiment---Effect of Two-Layer Predictor} 

\begin{figure}
\begin{center}
\includegraphics[width=0.7\columnwidth]{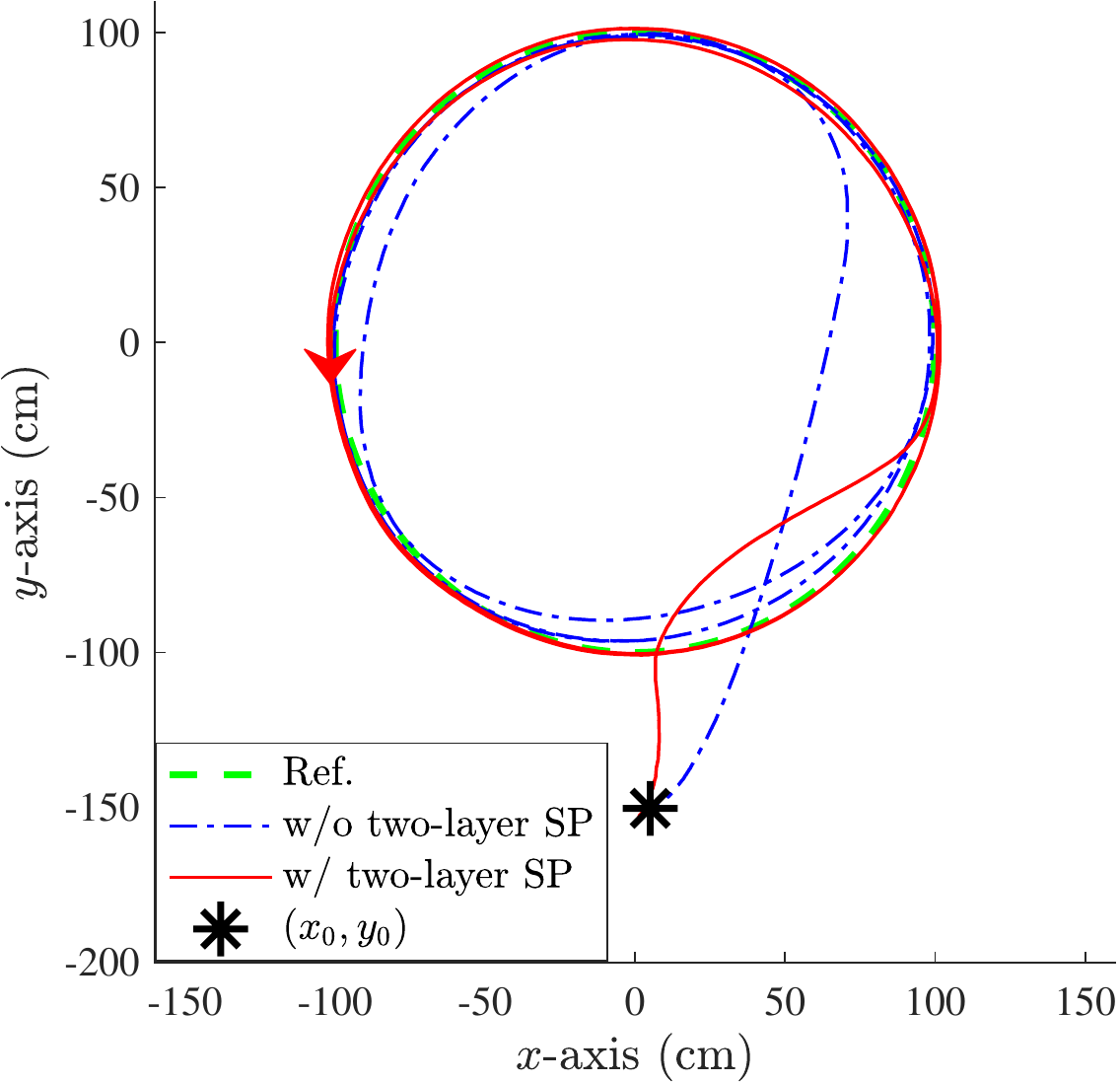}
\end{center}
\vspace{-5mm}
\caption{Trajectory-tracking of a circular reference path. Experimental performance with (in solid red) and without (in dash-dotted blue) a two-layer Smith predictor.}
\label{fig:xySPnoSP}
\end{figure}
Numerous experiments are carried out to evaluate the effectiveness of the two-layer predictor to improve trajectory tracking. Control calibration is done empirically. Each wheel's servo-system is controlled using a PI control, i.e., $C(s)=2+1/s$. Moreover, the angle control is also designed as a PI, i.e., $C_\theta(s)=0.6+0.1/s$. The integral action noticeably reduces the steady-state tracking error. 

\begin{figure}
\begin{center}
\includegraphics[width=0.8\columnwidth]{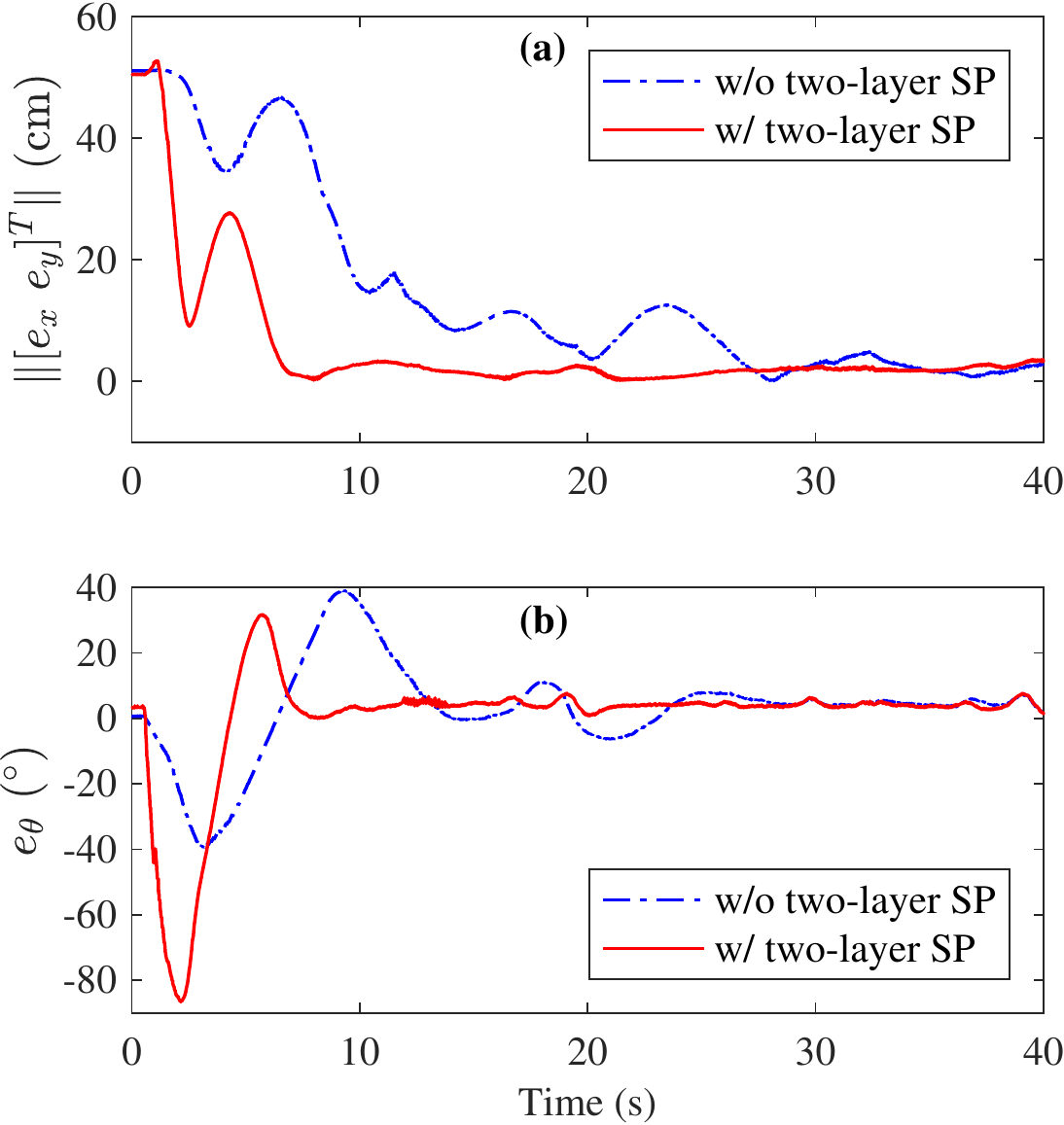}
\end{center}
\vspace{-5mm}
\caption{Trajectory-tracking of a circular reference path. Variation of (a) position error and (b) angle error versus time, with (in solid red) and  without (in dash-dotted blue) the two-layer Smith predictor. With the two-layer predictor, the transient is passed in less than $7$~s.}
\label{fig:errorSPnoSP}
\end{figure}

First, a circular reference trajectory is generated as $x_r=R\sin(\omega_r t), y_r=-R\cos(\omega_r t)$, where $R=1$~m, and $\omega_r=2\pi/20$~rad/s. The VFO without the two-layer predictor is unstable with $C(s)=2+1/s$. Thus, the servo-system control is modified as $C(s)=0.5+0.1/s$ when the two-layer predictor is not present. The initial condition is set to $x(0)=0.05$~m, $y(0)=-1.50$~m, and $\theta(0)=-3^\circ$. The effect of the two-layer Smith predictor (SP) is shown in Fig.~\ref{fig:xySPnoSP} and \ref{fig:errorSPnoSP}. Despite the large initial error, the proposed algorithm brings the robot to steady-sate in less than $7$~s, which improves the convergence time of the VFO by a factor of four. Contour error is defined as the closest distance from the actual position to the reference curve, directly measuring the tracking precision. The RMS and average value of the steady-state contour error for the proposed algorithm are $1.69$~cm and $1.57$~cm, respectively. 

\begin{figure}
\begin{center}
\includegraphics[width=0.99\columnwidth]{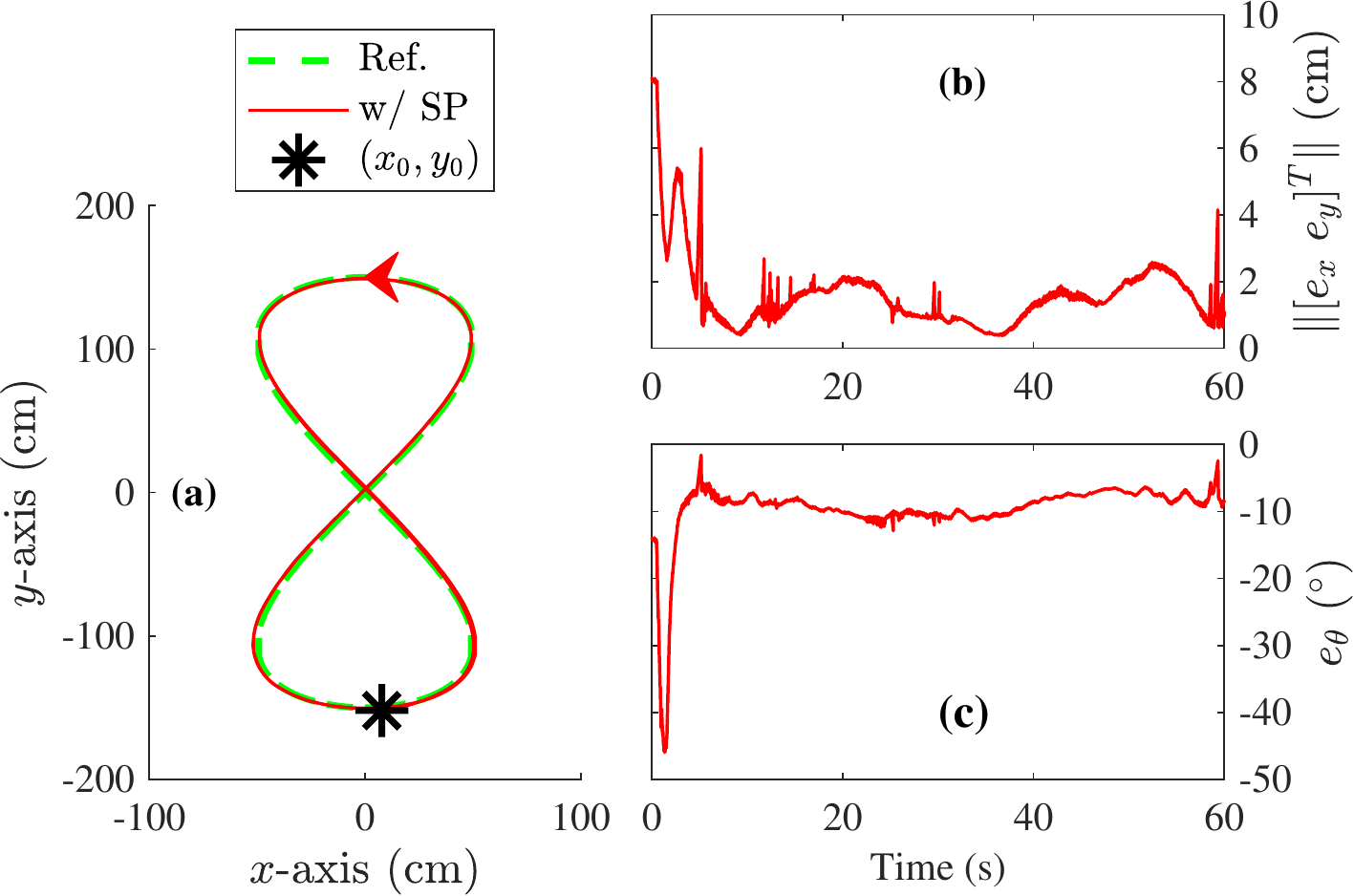}
\end{center}
\vspace{-5mm}
\caption{Performance of the proposed method for a figure-8 reference path. (a) Robot accurately tracks the reference trajectory. (b) Variation of position error versus time. (c) Variation of angle error versus time. The transient is passed in less than $5$~s.}
\label{fig:fig8}
\end{figure}
Moreover, the proposed method is tested with a figure-8 reference trajectory as given by $x_r=a_x\sin(2\omega_r t)$ and $y_r=-a_y\cos(\omega_r t)$, where $a_x=0.5$~m and $a_y=1.5$~m, and $\omega_r=2\pi/30$~rad/s. The initial condition is set to $x(0)=0.08$~m, $y(0)=-1.52$~m, and $\theta(0)=14~^\circ$. The experimental result is shown in Fig.~\ref{fig:fig8}. The algorithm passes the transient in less than $5$~s. The RMS and mean contour error are $1.28$~cm and $1.16$~cm.

\begin{remark}
The PI controller has limited ability to maintain steady-state angle error below a desirable range. If the reference angle exhibits complicated behavior, the steady-state error may increase. For example, in the circular reference, the angle increases with a fixed ramp. Thus, as shown in Fig.~\ref{fig:errorSPnoSP}(b), the PI controller keeps the RMS value of the steady-state angle error about $4^\circ$. However, the reference angle of the figure-8 oscillates between $\pm2.13$~rad. Thus, as shown in Fig.~\ref{fig:fig8}(c), the RMS value of the steady-state angle error is about $10^\circ$. Advanced angle control techniques require special investigation, which is not in the scope of this work. 
\end{remark}

\subsection{Second Experiment---Avoidance Maneuver with Multiple Circular Obstacles}

\begin{figure}
\begin{center}
\includegraphics[width=0.7\columnwidth]{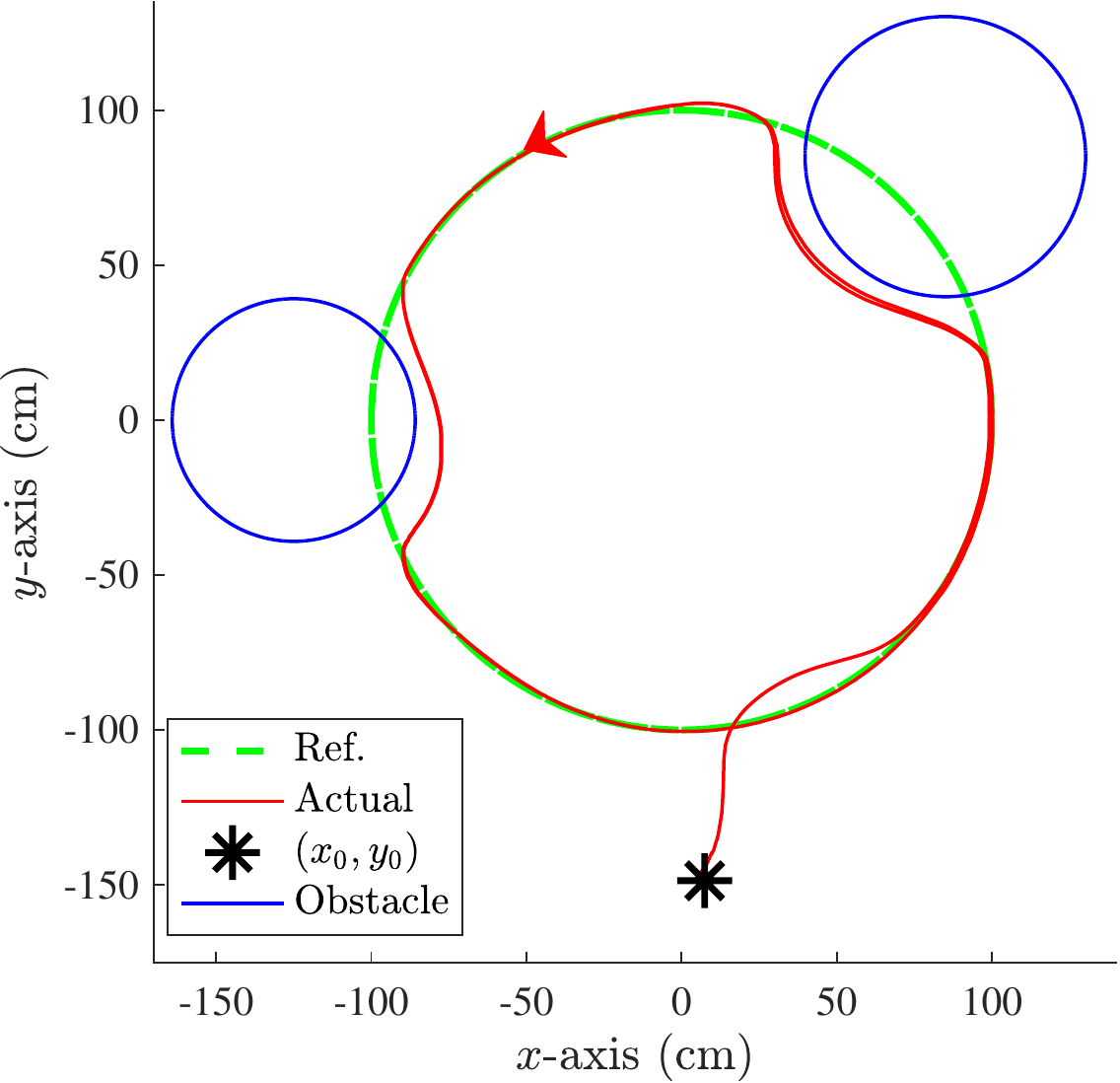}
\end{center}
\vspace{-5mm}
\caption{Experimental validation with two circular obstacles (in solid blue) on a reference circular path (in dashed-green). The QBot~2e consistently avoids the obstacles over multiple revolutions.}
\label{fig:CircularObstacles}
\end{figure}
\begin{figure}
\begin{center}
\includegraphics[width=0.8\columnwidth]{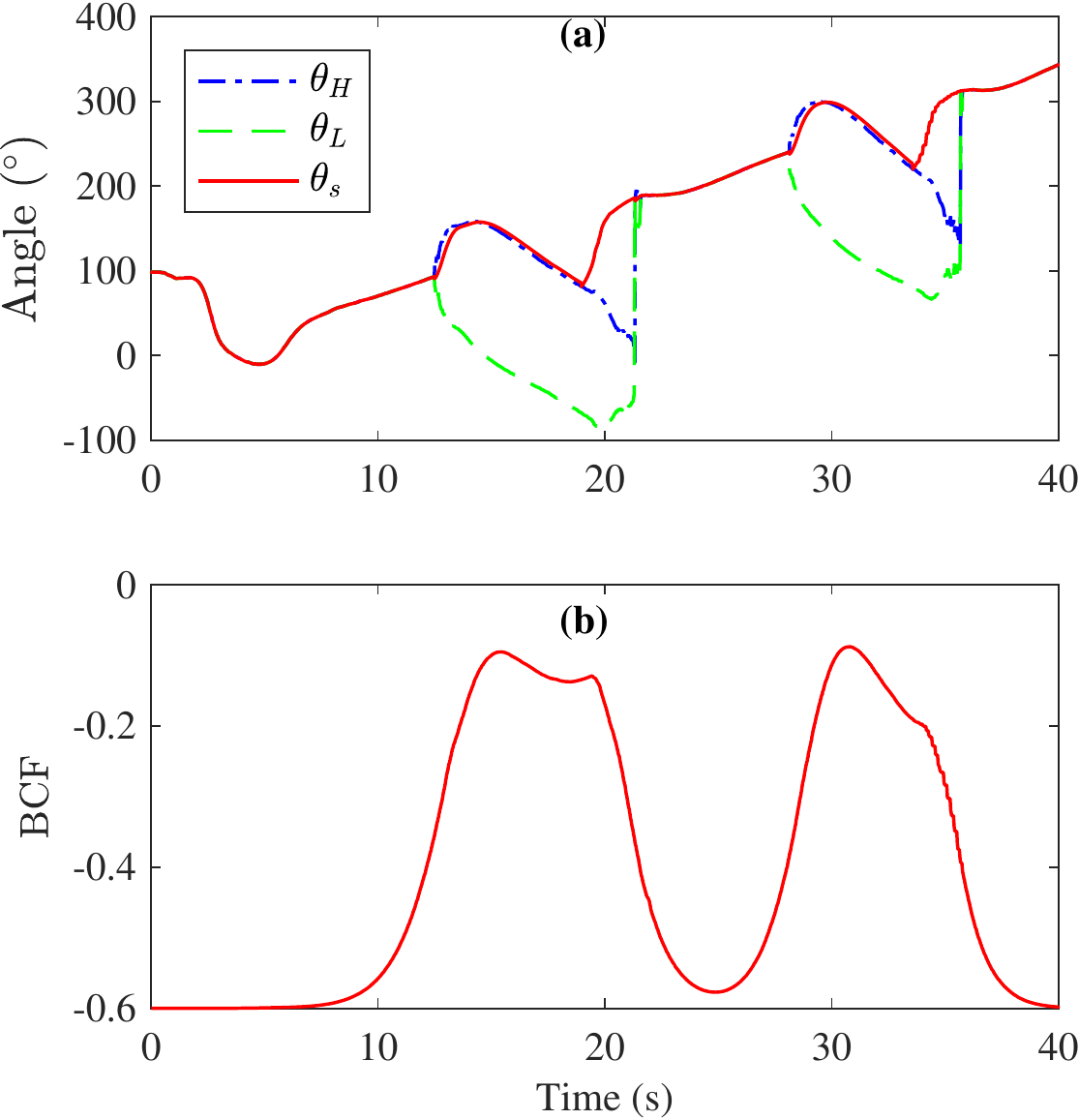}
\end{center}
\vspace{-5mm}
\caption{Experimental validation of safety-control of the QBot~2e with two circular obstacles. (a) The unsafe angle range is bounded between $\theta_L$ and $\theta_H$. The algorithm keeps the heading angle outside the unsafe angle range. (b) Negative value of $B(x,y)$ shows that the robot stays away from the obstacles.}
\label{fig:CircObsAng}
\end{figure}
The proposed algorithm can handle single or multiple obstacles in the operating environment. The control structure, including the VFO, the two-layer predictor, and the safe angle estimate, remains the same. The barrier certificate can also be formed using a general methodology, where the number, position, and dimension of the obstacles can be modified arbitrarily. For example, consider the following barrier certificate function (BCF)
\begin{equation}
\label{eq:Bexp}
B(x,y)=-B_0+\sum_{j=1}^{m}\exp\left(-{d_j^2}/{\sigma_j}\right),
\end{equation}
where $B_0$ and $\sigma_j$ are positive constants, $m$ is the number of obstacles, and $d_j$ is the distance of the robot from the obstacle $j$ calculated as
\begin{equation}
d_j=\sqrt{\left(x-x_{oj}\right)^2+\left(y-y_{oj}\right)^2},
\end{equation}
where $[x_{oj}~~y_{oj}]^T$ is the position of obstacle $j$ and $[x~~y]^T$ is the position of the robot. 

An appropriate selection of $B_0$ and $\sigma_j$ can model arbitrary avoidance radii for all the obstacles. For example, two obstacles with different avoidance radii are considered for this experiment. The obstacles are located at $[0.85~~0.85]^T$~m and $[-1.25~~0]^T$~m, where $\sigma_1=0.4$, $\sigma_2=0.3$, and $B_0=0.6$. The designer is free to choose any barrier certificate function for the obstacle avoidance as long as the conditions~\eqref{eq:B1}--\eqref{eq:B3} are satisfied. 

\begin{figure}
\begin{center}
\includegraphics[width=0.8\columnwidth]{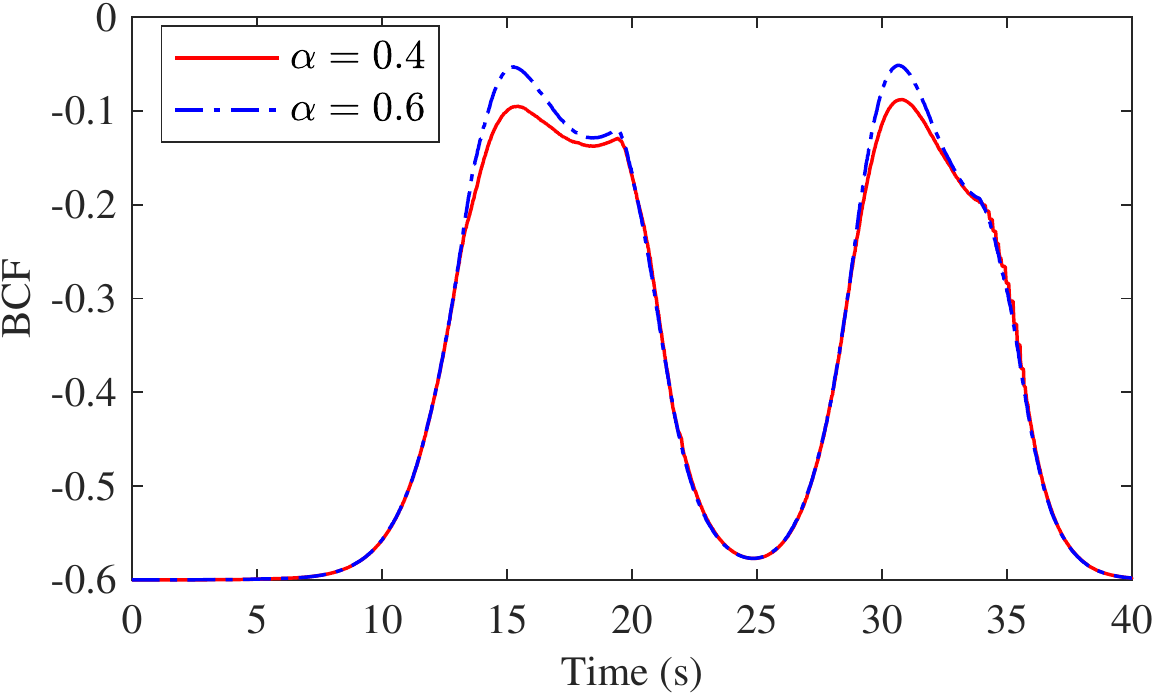}
\end{center}
\vspace{-5mm}
\caption{Effect of parameter $\alpha$ on the avoidance maneuver. As $\alpha$ increases, the robot moves closer to the avoidance radius. Thus, $B(x,y)$ can experience values close to zero. The safety-control may fail if $\alpha$ is increased beyond a certain level.}
\label{fig:CircObsB}
\end{figure}
The experimental result is shown in Fig.~\ref{fig:CircularObstacles} and \ref{fig:CircObsAng}. The reference trajectory is generated as $x_r=R\sin(\omega_r t), y_r=-R\cos(\omega_r t)$, where $R=1$~m, and $\omega_r=2\pi/40$~rad/s. The initial position and heading angle are $x(0)=0.07$~m, $y(0)=-1.48$~m, and $\theta(0)=3^\circ$. The algorithm provides consistent avoidance performance. The robot stays away from the obstacles shown as blue circles. Additional obstacles can be included by properly modifying the barrier certificate function~\eqref{eq:Bexp}. The calculated range of the unsafe heading angle is shown in Fig.~\ref{fig:CircObsAng}(a), where the adjusted heading angle, $\theta_a$, is not allowed to take any value between $\theta_L=\beta-\delta$ and $\theta_H=\beta+\delta$. In other words, the safe angle is kept outside the unsafe range during the experiment. As Fig.~\ref{fig:CircObsAng}(b) shows, the barrier certificate function stays below zero, which proves that safety is achieved.

The selection of parameter $\alpha$ in \eqref{eq:B3} affects the obstacle avoidance maneuver. Small values of $\alpha$ cause a conservative obstacle avoidance, which creates a long detour around the obstacle. On the other hand, a large value of $\alpha$ causes the corresponding value of $\delta$ to remain near zero, which means the safe angle estimate is not accurate. Thus, large values of $\alpha$ may cause the robot to collide with the obstacle. The experimental result shown in Fig.~\ref{fig:CircObsB} verifies the effect of $\alpha$ on the variation of the barrier certificate function. If one increases the value of $\alpha$, the robot may narrowly evade the obstacle. Hence, the barrier certificate experiences values near zero, as shown in Fig.~\ref{fig:CircObsB}. Thus, the upper limit of $\alpha$ needs to be selected carefully. 

\subsection{Third Experiment---Non-Circular Obstacles}

\begin{figure}
\begin{center}
\includegraphics[width=0.7\columnwidth]{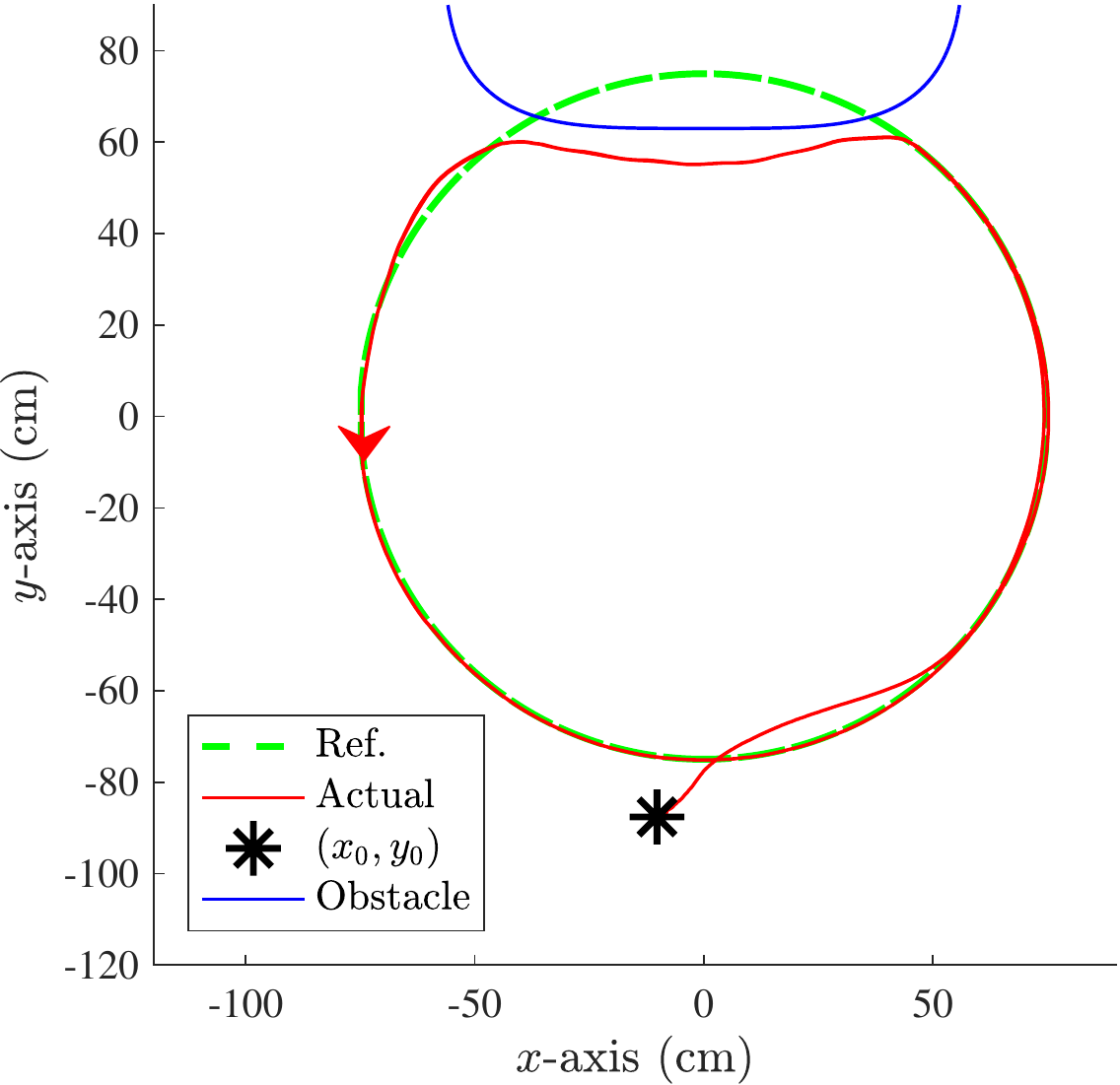}
\end{center}
\vspace{-5mm}
\caption{Experimental validation with a square obstacle (in solid blue) on a reference circular path (in dashed green). The QBot~2e consistently avoids the obstacles over multiple revolutions.}
\label{fig:SquareObstacle}
\end{figure}

\begin{figure}
\begin{center}
\includegraphics[width=0.8\columnwidth]{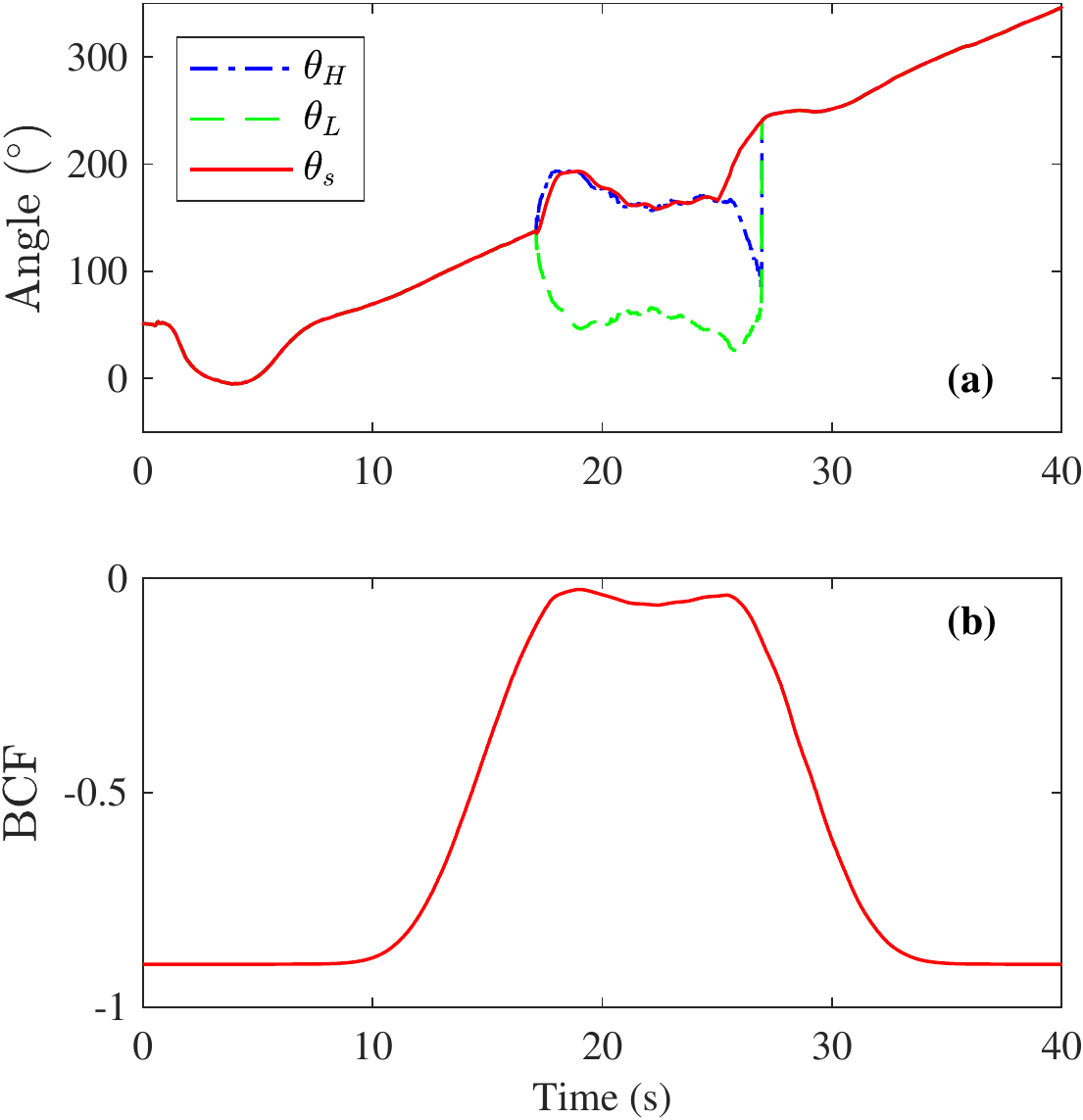}
\end{center}
\vspace{-5mm}
\caption{Experimental validation of safety-control of the QBot~2e with a square obstacle. (a) The heading angle is kept outside the unsafe range specified by $\theta_L$ and $\theta_H$. (b) The barrier certificate stays negative. Hence, the robot does not collide with the obstacle.}
\label{fig:SquareObsAng}
\end{figure}

The proposed safety algorithm can also handle non-circular obstacles. The components of the algorithm remain the same. A modified barrier certificate is needed to model the obstacle properly. For example, a square obstacle can be expressed as
\begin{equation}
B(x,y)=-B_0+\exp\left(-\left(\frac{x-x_o}{\sigma_x}\right)^{2n}\!\!-\!\!\left(\frac{y-y_o}{\sigma_y}\right)^{2n}\right),
\end{equation}
where $n$ is positive integer larger than one, $B_0$ is a positive real number, $\sigma_x$ and $\sigma_y$ are positive real numbers which specify the dimensions of the obstacle along $x$- and $y$-axis, respectively. Note that $n=1$ models an ellipse. Additional circular or square obstacles can be modeled by adding similar exponential terms with appropriate values for $\sigma_x$, $\sigma_y$, and $n$ for each obstacle. Here, a square obstacle is considered at $[0~~1.2]^T$~m with $\sigma_x=\sigma_y=1$ and $\alpha=1$. A circular reference trajectory is generated as $x_r=R\sin(\omega_r t), y_r=-R\cos(\omega_r t)$, where $R=0.75$~m, and $\omega_r=2\pi/40$~rad/s. The initial condition is set to $x(0)=-0.1$~m, $y(0)=-0.87$~m, and $\theta(0)=2^\circ$. The experimental results are shown in Fig.~\ref{fig:SquareObstacle} and \ref{fig:SquareObsAng}. The heading angle is kept outside the unsafe angle range. Thus, the robot successfully avoids the obstacle. When the robot is enough far from the obstacle, the trajectory tracking is accurate. Also, the barrier certificate function is kept below zero during the experiment.

\section{Conclusions}\label{sec:con}

A scalable and modular trajectory tracking with a two-layer predictor and integrated barrier certificates for obstacle avoidance is proposed for the safe and agile operation of nonholonomic mobile robots with time-delay. The two-layer predictor dramatically improves the transient performance of the heading angle control and servo-system loops. Barrier certificate functions are used to maintain a safe heading angle and thus avoid obstacles. The barrier certificate can model an arbitrary number of obstacles with different footprints. The structure of the proposed control is fully modular and independent of the obstacle properties. Thus, the proposed algorithm guarantees precision tracking and fast, successful avoidance maneuvers. Hence, the control design is dramatically simplified. The safety-control has been implemented in the form of an independent adaptive saturation block determining the robot's allowable heading angle, which steers the mobile robot clear of in-path obstacles. This block does not interfere with other components of the control system and hence makes the design modular and compatible with a wide class of control architectures. The conducted experiments verify that desirable closed-loop performance is achieved using linear control components. Future work addresses the extension of the proposed scalable and modular control architecture to provide safety-control of mobile robots in uncontrolled environments in the presence of moving obstacles.

\bibliographystyle{IEEEtran}


\end{document}